\documentclass[letterpaper, 10 pt, conference]{ieeeconf}  

\IEEEoverridecommandlockouts                              

\usepackage{amsmath} 
\usepackage{xcolor}
\usepackage{svg}
 \usepackage{multirow} 
 \usepackage{diagbox} 

\usepackage{graphicx}
\usepackage{amssymb}
\newcommand{\rev}[2]{{\color{black}{#2}}}

\title{\LARGE \bf Learning Dynamics of a Ball with  Differentiable Factor Graph and Roto-Translational Invariant Representations}

\author{Qingyu Xiao$^{1}$,  Zixuan Wu$^{1}$ and Matthew Gombolay$^{1}$
\thanks{This work was supported by Google under Award 004686. }
\thanks{$^{1}$ The authors are affiliated with the Georgia Institute of Technology, Atlanta, GA 30332 USA.}
\thanks{Corresponding author: Qingyu Xiao (qxiao33@gatech.edu)}
}

\begin{document}

\maketitle
\thispagestyle{empty}
\pagestyle{empty}

\begin{abstract}    
\rev{*}{
Robots in dynamic environments need fast, accurate models of how objects move in their environments to support agile planning.
In sports such as ping pong, analytical models often struggle to accurately predict ball trajectories with spins due to complex aerodynamics, elastic behaviors, and the challenges of modeling sliding and rolling friction. 
On the other hand, despite the promise of data-driven methods, machine learning struggles to make accurate, consistent predictions without precise input.
In this paper, we propose an end-to-end learning framework that can jointly train a dynamics model and a factor graph estimator.
Our approach leverages a Gram-Schmidt (GS) process to extract roto-translational invariant representations to improve the model performance, which can further reduce the validation error compared to  data augmentation method. Additionally, we propose a network architecture that enhances nonlinearity by using self-multiplicative bypasses in the layer connections. By leveraging these novel methods, our proposed approach predicts the ball's position with an RMSE of 37.2 mm of the paddle radius at the apex after the first bounce, and 71.5 mm after the second bounce. 
}

\end{abstract}
\section{INTRODUCTION}
Agile robotics generally operate in fast and dynamic environments, where the ability to predict future states of the environment is crucial for  navigation and planning. In applications ranging from autonomous driving \cite{song2020pip} and human-robot collaboration \cite{mainprice2013human} to competitive sports \cite{zaidi2023athletic} and various other fields \cite{zhu2021learning, aghili2012prediction}, the development of real-time and accurate predictive models is essential. In sports such as ping pong or tennis, developing a robotic partner capable of playing with humans \cite{lee_krishna_2023} presents unique challenges. Fine-tuning analytical models or training neural networks with real-world data is difficult because the states, such as position, velocity, and spin, are often noisy or unobservable. In this paper, we bridge the data-drive and optimization methods to learn roto-translational invariant models in a sample-efficient manner.

\begin{figure}
    \centering
    \includegraphics[width=0.99\linewidth]{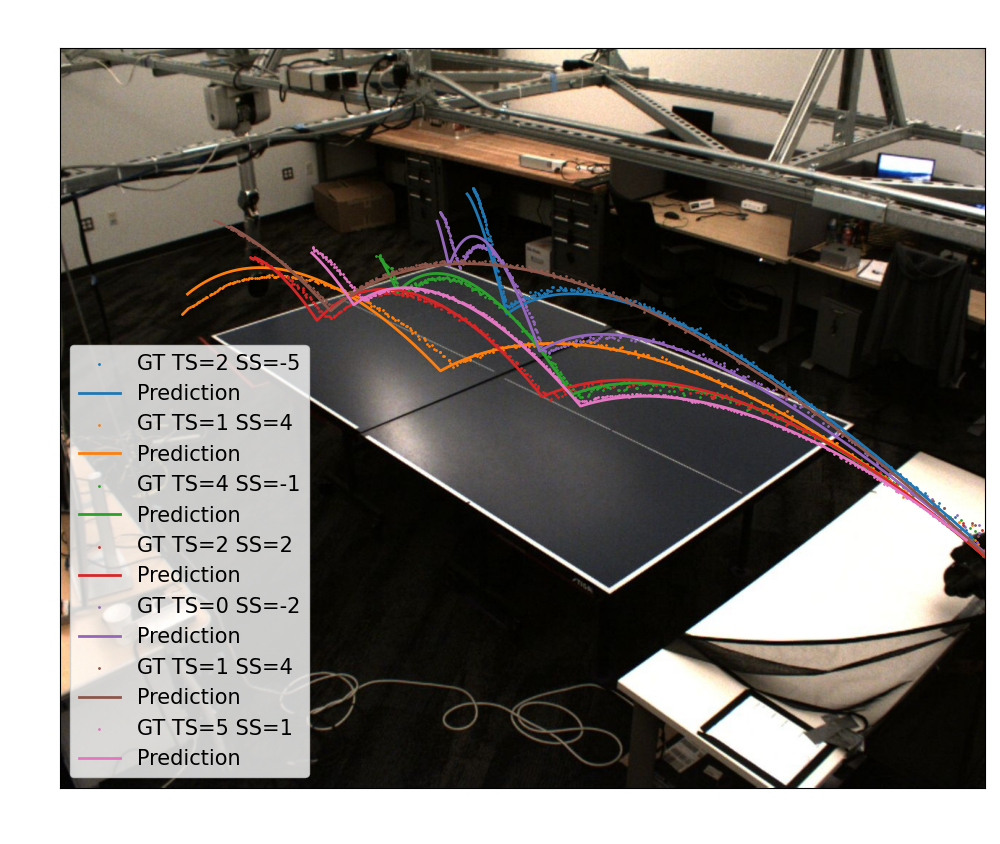}
    \caption{Predicted ball trajectories conditioned on the launcher's spin settings. Spin values are represented as integers, where larger positive numbers or smaller negative numbers indicate higher spin. In the legend, 'TS' denotes topspin and 'SS' denotes sidespin.}
    \label{fig: predicton}
\end{figure}

\rev{*}{Recent efforts have been made toward developing robotic systems for ping pong  partners \cite{ding2022goalseye, ding2022learning, d2023robotic}, resulting in the creation of an amateur-level ping pong robot \cite{d2024achieving}. However, this work also highlights the need for further improvements due to inaccuracies in state estimation and trajectory prediction, particularly for balls with strong spin adding magnus effects \cite{d2024achieving}. Some works \cite{zhang2014spin, gossard2024table} attempt to measure spin directly by tracking logos or markers on the ball, but the resulting measurements are often noisy and unreliable for accurate dynamics learning and prediction. A promising alternative is to use a professional ball launcher, which can precisely control spin by adjusting the launcher settings \cite{achterhold2023black}. Although the launcher settings do not provide exact spin measurements in revolution rates, they offer reliable indications of spin strength. Our approach leverages this method by utilizing the launcher settings for trajectory labeling as shown in Figure \ref{fig: predicton}.}


Many studies \cite{zhang2014spin, zhang2015real, zhao2017model} rely on analytical models, including ball aerodynamics \cite{meng1996mechanical} and bounce dynamics \cite{bao2012bouncing}. However, these models are derived under assumptions that do not hold in real-world scenarios and lead to significant prediction errors. Data-driven approaches, such as LSTMs \cite{hochreiter1997long}, diffusions \cite{janner2022planningdiffusionflexiblebehavior, 10416775} and generative models \cite{gomez2020real}, use raw data observations directly as input. Similarly, other work \cite{allen2023graph} utilizes a multilayer perceptron to learn dynamics, representing states through finite difference velocity derived from raw data. However, these methods learned from real world data often suffer from the lack of precise initial state information, which significantly affects the accuracy of long-term predictions. Estimators such as kalman filters (EKF \cite{ribeiro2004kalman} or AEKF \cite{jetto1999development}) or factor graph based method can provide better initial state estimation. However, their performance is closely tied to the accuracy of the dynamics model used for inference.  This  motivates our work to develop an end-to-end learning framework that can jointly learns both the estimator and the dynamics model.

In this paper, we improve trajectory prediction for balls with various types of spin using an end-to-end dynamics learning framework. The key contributions include: proposing an end-to-end learning framework that jointly learns the estimator and dynamics model using a differentiable factor graph; employing the Gram-Schmidt process in the dynamics model to extract roto-translational invariant representations; and developing a neural network with a self-multiplicative bypass to further improve prediction accuracy. 

\section{RELATED WORK}

\subsection{Estimator}
\label{subsec: Estimator}
 Estimators play a crucial role in trajectory prediction, as they rely heavily on accurate initial state estimation. \rev{*}{For pingpong trajectory prediction, the Kalman filters, such as extended kalman filter (EKF) or adaptive EKF (AEKF) \cite{jetto1999development}, is commonly employed to estimate position and velocity \cite{d2023robotic,zhang2014spin,gomez2020real}.} However, the kalman filters, viewed as a specialized chain-based factor graph, processes estimations in a sequential manner, discarding early observations which can lead to significant information loss compared to optimizing an entire factor graph \cite{dellaert2017factor}. Recent studies have demonstrated that factor graphs yield better state estimation for tennis balls compared to Kalman filters \cite{xiao2024multi}. These findings are relevant given the similarities between tennis and pingpong balls. Recent advancements in differentiable factor graph estimators \cite{yi2021differentiable, wang2023pypose, pineda2022theseus} enable end-to-end learning through optimization. Consequently, utilizing the differentiable factor graph enables the joint learning of both the estimator and the dynamics model.

\subsection{Spin}
Many works have attempted to directly measure the spin by detecting logos or markers on the ball \cite{zhang2015real, tamaki2024spin, glover2014tracking}. These methods require cameras to be positioned close enough to capture clear views of the logos and rely on high-speed cameras to match the frequency of the ball's spin. However, even when these challenges are addressed, the estimation remains noisy due to the limited number of observations within a short time window \cite{gossard2024table}. A promising approach to labeling the initial spin of the ball is to use the settings from a launcher, where the spin can be computed based on the motorized wheels of the launcher, which have measurable spin rates for each wheel \cite{achterhold2023black}. In our approach, we adopt a similar idea, but instead of computing the spin from the wheels to the ball, we directly label the trajectory using the spin indications from the launcher settings based on the launcher settings to avoid extra equipment or computational errors for spins.

\subsection{Ball Dynamics Model}
Analytical models for pingpong dynamics have been derived, including those for aerodynamics \cite{meng1996mechanical} and bounce dynamics \cite{bao2012bouncing}. Using these analytical models, the ball's trajectory can be predicted by numerically integrating the ordinary differential equations (ODEs) established from the derived equations. Analytical models are widely used in current literature \cite{zhang2014spin, zhang2015real, zhao2017model}. Therefore, we will include analytical models and fine tune the parameters on our dataset as one of our benchmarks. Learning-based approaches have also been explored, such as LSTM and conditional generative models \cite{gomez2020real}, showing significant improvement over fine-tuned analytical models. However, this prior work does not account for ball trajectories involving spins. 
In our work, we will compare our proposed method with LSTM and generative models as additional benchmarks.

\section{DYNAMICS LEARNING FRAMEWORK}

Dynamics describe the evolution of an object's motion over time based on its initial conditions and the forces acting upon it. Even a perfect dynamics model will lead to large error in prediction if the initial state is estimated inaccurately. Learning the dynamics of a ball from real world data reliable initial state or initial state representations. Due to the noise in the position data and the challenge of measuring the ball's velocity and spin, estimators are need to refine these initial states. However, since estimators need accurate model for precise estimation, accurate models also need precise estimator so that the model can learn from real world data. This creates a chicken-and-egg problem, where both the model and the estimator depend on each other.

\begin{figure*}[h]
    \centering
    \includegraphics[width=0.70\linewidth]{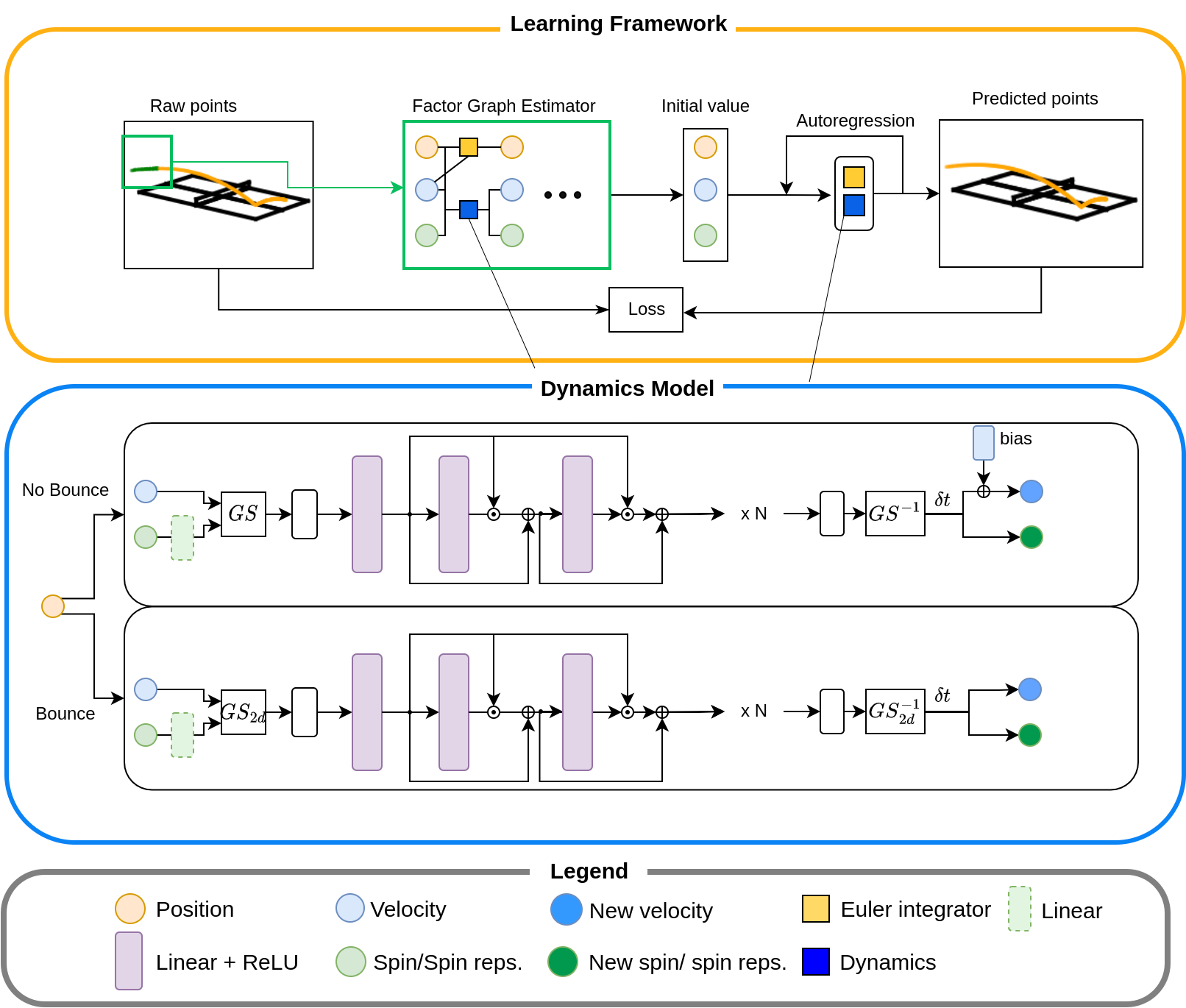}
    \caption{The end-to-end dynamics learning framework and the proposed self-multiplicative neural netowrk (MNN) architecture for dynamics learning.}
    \label{fig: dynamics learning framework}
\end{figure*}

To address this, we propose an end-to-end learning framework based on a differentiable factor graph. Figure \ref{fig: dynamics learning framework} presents an overview of the proposed learning framework, which consists of two main components: (a) a factor graph estimator to find the optimal initial values, as discussed in Section \ref{subsec: factor graph estimator}, and (b) autoregressive prediction using a dynamics model, detailed in Section \ref{subsec: dynamics model}. The framework minimizes the prediction error across the entire trajectory, influenced by both the factor graph and the dynamics model. This loss is then backpropagated through the network to update the parameters of both the dynamics model and the differentiable estimator.

\subsection{Factor Graph Estimator}
\label{subsec: factor graph estimator}
Before predicting the trajectory, we build factor graph based on the first $N_{obs}$  observations and and any priors to estimate the initial states of the ball. Due to the nonlinear nature of the dynamics model, we need nonlinear optimizers such as the Levenberg-Marquardt (LM) algorithm \cite{ranganathan2004levenberg} to optimize the factor graph likelihood. The advantage of using a differentiable factor graph is that it can automatically compute the Jacobian of the factors (i.e., the dynamics model) required by the LM algorithm. This eliminates the need to manually compute the Jacobian when modifying the model architecture or updating weight parameters during training. Additionally, differentiable factor graph can also learn the noise parameters of the estimator if the noise cannot be modeled properly.

The entire factor graph can be factorized as shown in  (\ref{eq: graph factorization}), where $\eta$ denotes all the variables in the graph, $f_i$ represents the potential function of the $i^{\text{th}}$ factor, $\xi_i$ refers to all the variables associated with the $i^{\text{th}}$ factor, and $M$ is the total number of factors in the graph.

\begin{equation} G(\eta) = \prod_{i=0}^{M-1} f_i(\xi_i) \label{eq: graph factorization} \end{equation}

The potential function for each factor is expressed in  (\ref{eq: factor}):

\begin{equation} f_i(\xi_i) = \exp\left(-\frac{1}{2}|err_i(\xi_i)|^2_{\Sigma_i}\right) \label{eq: factor} \end{equation}

Here, $err_i$ is the error function for the $i^{\text{th}}$ factor, and $\Sigma_i$ is the covariance matrix corresponding to that factor. Thus, factor graph optimization is equivalent to solving a system of linear equations, with the weight matrix given by $\Sigma_i^{-1}$. Many differentiable factor graphs library are available \cite{yi2021differentiable, wang2023pypose}. In this paper, we build our factor graph and convert it to a nonlinear system which can be optimized in Theseus \cite{pineda2022theseus} to enable batch learning.

\subsection{Dynamics Model Details}
\label{subsec: dynamics model}

In this section, we present the proposed method, which involves: (1) using the Gram-Schmidt process to extract roto-translational invariant representations, and (2) employing a self-multiplying neural network, which add additional multiplication bypass in the hidden state, to enhance the network's nonlinearity.

\subsubsection{Roto-translational invariant representations}
\label{subsec: roto-trans invar}
\rev{*}{A roto-translational invariant representation refers to a feature representation that remains unchanged when the input data undergoes rotation or translation \cite{kofinas2021roto}.} From physics point of view, the force acting on the ball follows this roto-translational invariant features. In order to enable the machine leanring method to strictly follows this features, we propose to use Gram-Schmidt (GS) vectorization method to extract roto-translational invariant representations. The GS vectorization can be written as (\ref{eq: gs vectorization}).
\begin{equation}
    \mathbf{q}_k = \mathbf{x}_k - \sum_{j=1}^{k-1} \mathbf{q}_j^T \mathbf{x}_k  \frac{ \mathbf{q}_j}{\mathbf{q}_j^T \mathbf{q}_j}
    \label{eq: gs vectorization}
\end{equation}

We define the representations extracted from $n$ input vectors $X = \{\mathbf{x}_j\}_{j<n}$ using GS method as
\begin{equation}
GS(X) = \left\{ \frac{\mathbf{x}_j^T \mathbf{q}_k}{\sqrt{\mathbf{q}_k^T \mathbf{q}_k}} \right\}_{0 \leq k, j < n}
\end{equation}

 Given an arbitray rotation matrix $R$ that rotates each vector $\mathbf{x}_k$ by the matrix . The rotated vectors are $R \mathbf{x}_k$. The Gram-Schmidt process applied to the rotated vectors yields new basis $\mathbf{q}'$ where

\begin{eqnarray}
    \mathbf{q'}_1 &=& R \mathbf{x}_1\\
    \mathbf{q'}_k &=& R \mathbf{x}_k - \sum_{j=1}^{k-1} \frac{(R \mathbf{q}_j)^T (R\mathbf{x}_k)}{(R\mathbf{q}_j)^T (R\mathbf{q}_j)} R\mathbf{q}_j
\end{eqnarray}

This result in
\begin{equation}
     \mathbf{q'}_k = R \left[ \mathbf{x}_k - \sum_{j=1}^{k-1} \frac{ \mathbf{q}_j^T \mathbf{x}_k}{\mathbf{q}_j^T \mathbf{q}_j} \mathbf{q}_j \right] = R \mathbf{q}_k
     \label{eq: q=Rq'}
\end{equation}

Using the conclusion in (\ref{eq: q=Rq'}), we can derive
\begin{equation}
    \begin{aligned}
        GS(RX) &= \left\{ (R\mathbf{x}_j)^T (R\mathbf{q}_k) / \sqrt{R\mathbf{q}_k^T R\mathbf{q}_k} \right\}_{0 \leq j , k < n} \\
               &= \left\{ \mathbf{x}_j^T \mathbf{q}_k / \sqrt{\mathbf{q}_k^T \mathbf{q}_k} \right\}_{0 \leq j < k < n} \\
               &= GS(X)
    \end{aligned}
\end{equation}

Hence, the representation extracted from $GS(\cdot)$ is rotationally invariant. Note if $X$ does not have positions, the model will be roto-translational invariant since it will not be affected by position. In this paper, we use the GS method to extract roto-translational representations from translation-invariant states (in our case, velocity and spin), which are then fed into the neural network as hidden states. Typically, in aerodynamics learning, we apply the GS method in 3D space. However, in bounce dynamics learning, we use the GS method in 2D space (x- and y-axis, noted as $GS_{2d}$ in Figure \ref{fig: dynamics learning framework}), as the knowledge of bounce dynamics only preserves invariant representations when rotating along the z-axis.

\subsubsection{Spin Representations}
Ideal spin representations should have clear physical meaning, such as revolutions per minute (RPM) or Hertz. However, accurately measuring the revolution rate of a ball using standard cameras is challenging. One approach is to measure the revolution rate of the launcher wheels and compute the spin on the ball, though this can introduce errors from factors such as sliding friction between the wheels and ball, or inaccuracies in wheel radius measurements. Instead of relying on physics-based spin representations, our work focuses on spin representations derived from the launcher settings, ("TS" and "SS" as shown in Figure \ref{fig: predicton}), with more details provided in Section \ref{subsec: dataset}. Although the values of these launcher-derived representations are not intuitive in a physical sense, our work shows models trained using these representations have demonstrated strong performance in predicting the ball's trajectory.

\subsubsection{Self-Multiplying  Neural Network}
\label{subsec: self-multiplying NN}
In order to learn the  nonlinear dynamics of the ball, we propose the self-multiplying neural network (MNN). As shown in Figure \ref{fig: dynamics learning framework}, the initial spin representation is connected to a linear layer, which converts the discrete launch settings into a 3D vector that aligns with the velocity. However, if the RPM could be measured accurately and used as input, this layer is not necessary. Next, we apply the Gram-Schmidt (GS) method to extract roto-translational invariant representations and project them into hidden states $h_0$ via a linear layer. The linear layer $F(\cdot)$ with skipped conections and self-multiplicative connections repeatedly apply to the hidden states $N$ times.  This neural network features skip connections similar to ResNet \cite{he2016deep}, but with an added self-multiplying bypass to enable a higher degree of nonlinear behavior. The repeated $k^{\text{th}}$ process is formalized in  (\ref{eq: mnn iteration})

\begin{equation}
    \mathbf{h}_k = \mathbf{h}_0 F(\mathbf{h}_{k-1}) + \mathbf{h}_{k-1}
    \label{eq: mnn iteration}
\end{equation}

Since the hidden states $h_k$ are derived from roto-translational invariant representations, we apply the inverse of the Gram-Schmidt ($GS^{-1}$) to convert these representations back to the world coordinate system, yielding acceleration and angular acceleration in the world frame. 

Noting that slight difference between aerodynamics model and bounce dynamics model: 1) The bounce dynamics model uses the GS method in 2D space (x- and y-axis) because bounce dynamics only preserve rotationally invariant representations when rotating along the z-axis. 2) A bias is added to the last layer of the aerodynamics model to compensate for gravity. Therefore, while the aerodynamic force itself is rotationally invariant, the trajectory predicted by the aerodynamics model retains this rotational invariance only when rotated along the vector defined by the bias.

\section{RESULTS} 

\rev{*}{In this section, we discuss the domains and datasets used to validate our proposed approach. We outline the benchmarks used for comparison and describe the metrics employed for evaluation. Finally, we present our results, including both computational analysis and real-world data.
}

\subsection{Domains and Dataset}
\label{subsec: dataset}
\begin{figure}[h]
    \centering
    \includegraphics[width=0.99\linewidth]{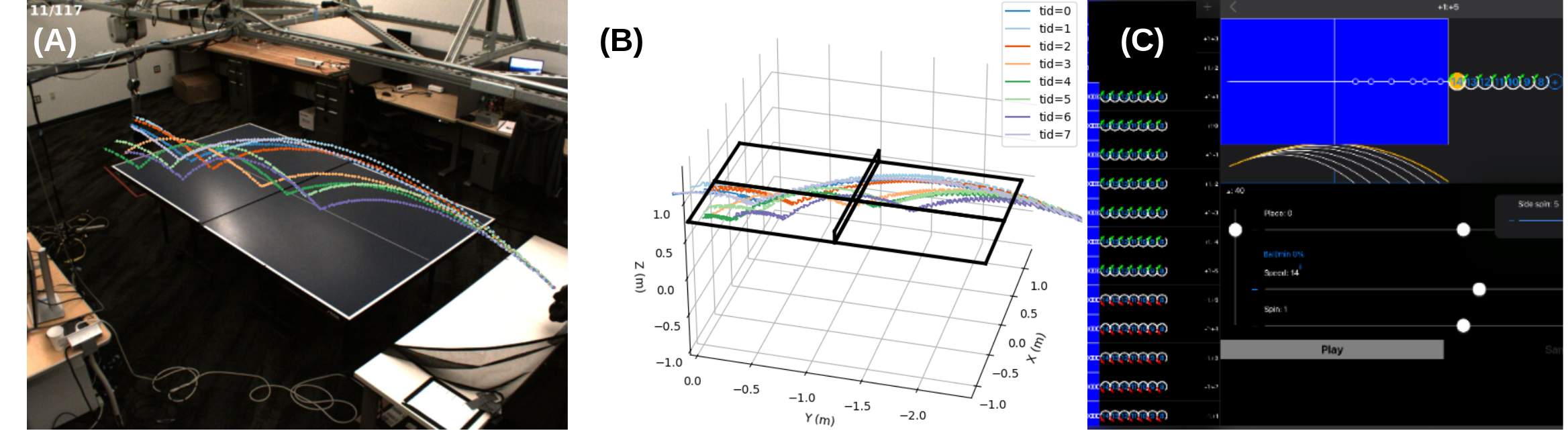}
    \caption{Experiment setups. (A) shows the view of one of the three calibrated cameras are deployed around the table. (B) visualizes the raw 3D points are computed by triangulation from paired cameras. (C) Depicts the spin indications from launcher settings.}
    \label{fig: setups}
\end{figure}

We evaluate our proposed approach in pingpong scenarios using a dataset recorded with three calibrated cameras positioned around the table. The view of one of the cameras is shown in Figure \ref{fig: setups}A. The 3D positions, shown in Figure \ref{fig: setups}B, are computed via triangulation \cite{hartley1997triangulation} using detections from a YOLO-LITE \cite{huang2018yolo} model fine-tuned on images of the ping pong balls. The Figure \ref{fig: setups}C illustrate the recorded trajectories are labeled based on launcher spin settings.
The dataset consists of 717 trajectories recorded under various launcher settings, including topspin levels in integers in [-3,5], sidespin levels in [-5,5], and velocity levels in [8,14]. Although the revolution rate of the spin cannot be measured directly, the effects offer insight into its magnitude. For example, a -3 topspin can cause the ball to bounce backward when velocity level below 10, while +5 or -5 sidespin can alter the ball's trajectory by approximately 45°. This indicates a wide range of spin intensities. Most trajectories involve a single bounce, but some include two or even three bounces.  Although multiple bounces are uncommon in typical pingpong gameplay (except for serves), these trajectories are critical for validating the dynamics model. 

\subsection{Benchmarks}
In this subsection, we introduce the benchmarks used for both state estimation and dynamics models.

\subsubsection{Estimators}

In this benchmark, we aim to demonstrate that the factor graph-based approach outperforms other methods, such as the sliding window mean and EKF, in estimating velocity under the same conditions. We generate 50 synthetic trajectories based on an analytical model with random initial values. Noise is added to these trajectories to simulate real-world conditions. The performance of each estimator is then compared based on velocity estimation accuracy

\begin{itemize}
    \item \textbf{Sliding Window Mean}: This method computes the average velocity within a sliding window of size 5.
    
    \item \textbf{EKF}: The EKF is a widely accepted Kalman filter commonly used for state estimation. The EKF is provided with the ground truth dynamics model and ground truth spin used to generate the trajectories. The velocity is estimated using EKF, with the initial state guess derived from the sliding window mean (window size of 5). The process noise (covariance) equals to the noise we added to the trajectory.
    
    \item \textbf{Factor Graph}: Factor graphs are well-known graphical models for state inference and are widely applied in robotics. The factor graph also has access to the ground truth dynamics model and spin. The initial guess is obtained in the same manner as the EKF, using the sliding window mean. Additionally, the same noise model (covariance) is applied to the factor graph for consistency.
\end{itemize}

\subsubsection{Dynamic models}
\rev{*}{
In this benchmark, we aim to demonstrate that our proposed learning framework achieves better prediction accuracy compared to other general-purpose neural networks. Additionally, we show that extracting roto-translational representations using GS, combined with the inclusion of a self-multiplicative layer in the hidden states, further enhances prediction accuracy. To this end, we compare the following dynamics models:
\begin{itemize}
   \item \textbf{LSTM + Aug.} : LSTM is a representative model  of recurrent neural networks. This network is trained with data augmentation - randomly rotate along z-axis or translate the trajectory in x- or y-axis in the dataset.
    \item \textbf{Diffusion + Aug.} Diffusion is a representative model of generative models. This network is trained with same data augmentation method as LSTM.
    \item \textbf{A-Tune + Aug.}: This model adopt the analytical model but with the parameters fine-tuned by real world data. The training is based on our end-to-end learning framework with a differentiable factor graph estimator. We apply random translation and z-axis rotation to the dataset to achieve better generalization of the model.
    \item \textbf{MLP + Aug.}: This model replace the analytical model used in \textit{A-tune} with a multilayer perceptron while remain other components unchanged. 
    \item \textbf{MLP + GS}: This model incorporates a GS process to extract roto-translational invariant representations, ensuring the symmetric properties of the forces acting on the ball. No data augmentation is applied.
    \item \textbf{Skip + GS}: This model add skipped connections to MLP. This model aim to compare if the neural network achitecture can further improve the prediction accuracy.
    \item \textbf{MNN + GS (ours)}: Our proposed method, incorporating a self-multiplication bypass mechanism aside from skipped connections. 
\end{itemize}
}

\subsection{Metrics}
\rev{*}{
To assess the overall accuracy of each dynamics model, we calculate the RMSE of the 3D points along the entire trajectory. Additionally, to evaluate prediction accuracy after each bounce, we compute the RMSE of the apex, defined as the highest point between bounces. Specifically, we examine the apex before the first bounce, between the first and second bounces, and between the second and third bounces. The apex before the first bounce reflects the model’s accuracy in predicting the ball’s flight, where aerodynamic forces are the primary influence. The apex between the first and second bounces measures the accuracy of the predicted velocity and spin post-bounce, while the apex between the second and third bounces indicates the model’s consistency in maintaining accurate predictions of velocity and spin throughout the trajectory.

}
\subsection{Results and Analysis}
\subsubsection{Estimators comparison}
\begin{figure} [h]
\centering
\includegraphics[width=0.99\linewidth]{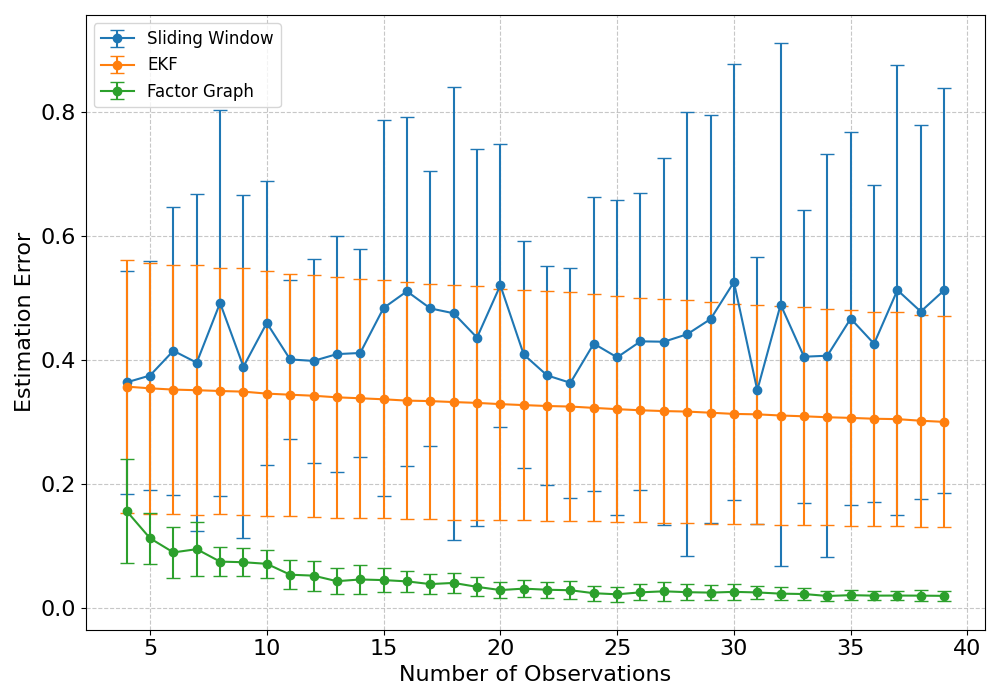} 
\caption{RMSE of estimated velocity using different estimators. EKF have better estimation compared to sliding window method when having more observations. But EKF have significantly larger estimation errors than factor graph-based estimator (used in our learning framework) in magnitude and standard deviation. } 
\label{fig: estimator compare} 
\end{figure}

The velocity estimation results are shown in Figure \ref{fig: estimator compare}. Both the EKF and factor graph estimators show decreasing estimation errors as the number of time steps increases. However, the factor graph estimator consistently achieves significantly lower estimation errors and smaller standard deviations compared to the EKF. These findings support the argument made in Section \ref{subsec: Estimator}, where the Kalman filter is described as a special case of the factor graph, with previous observations discarded \cite{dellaert2017factor}. Additionally, the results demonstrate that the factor graph estimator is more robust in handling noisy observations, providing more accurate state estimates over time, making it a better choice for initiating the dynamics learning process.

\subsubsection{Dynamics model comparison}
\begin{figure}[h]
    \centering
    \includegraphics[width=0.99\linewidth]{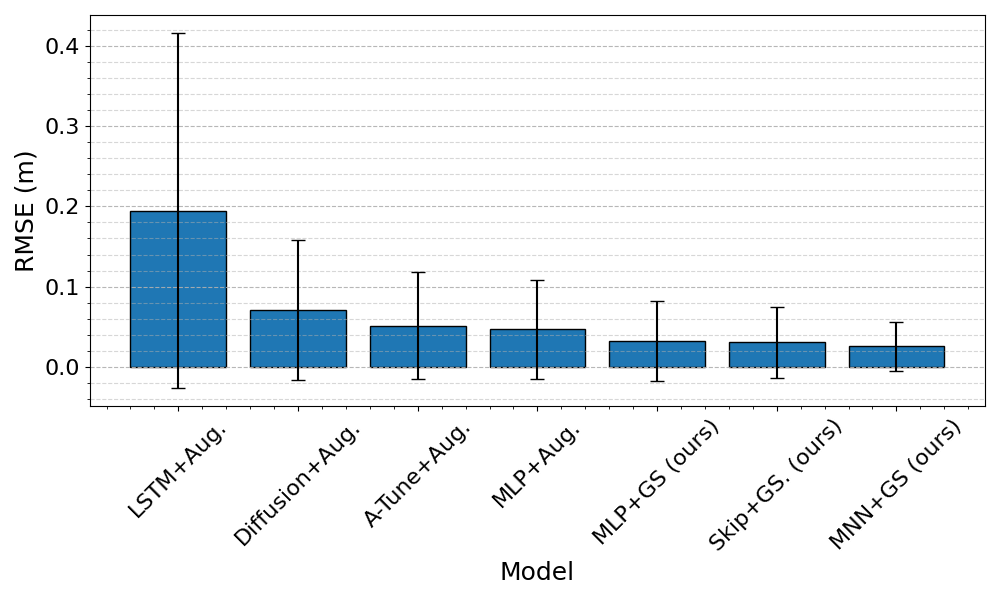}
    \caption{Compare the RMSE learned from different dynamics models. "Aug." means the model uses data augmentation instead of Gram Schmidt (noted as "GS") process to extract roto-invariant representations. "A-Tune" is analytical model with parameters learned from real world data. "Skip" refers to MLP with skipped connections, "MNN" refers to MLP with self-multiplying bypass along with skipped connections. }
    \label{fig: loss compare}
\end{figure}

The comparison of the validation loss from different dynamics models is shown in Figure \ref{fig: loss compare}. The A-Tune model (fine-tuned analytical model) and the diffusion model perform better than the LSTM, consistent with the findings in \cite{gomez2020real}. However, contrary to the conclusions in \cite{gomez2020real}, we find that A-Tune outperforms the generative diffusion model in our experiments.  We hypothesize this is because our dataset includes ball with spins, and A-Tune benefits from the differential factor graph in our proposed learning framework.

Our results demonstrate that using GS significantly improves prediction accuracy compared to standard augmentation techniques. The MLP model with augmentation achieves an RMSE of 0.0467 m, whereas the model with GS reduces the RMSE to 0.0322 m, marking a 31.0\% improvement. This improvement can be attributed to the roto-translational invariant representations used in the GS model, which are better suited for generalizing to real-world data.  

Additionally, the architecture of the model has a notable impact on errors. The MLP with GS achieves a RMSe of 0.0322 m, but by incorporating skip connections, the model's performance improves, achieving a loss of 0.0304 m,  an improvement of 5.59 \%. Finally, by adding self-multiplicative bypasses to introduce a higher degree of nonlinearity, we achieve a RMSE of 0.0253 m, representing an additional improvement of 16.8\% over the skip connection model.

\subsubsection{Apex Errors Analysis}
The results in Table \ref{tab: apex rmse} show that neural networks trained without the GS process exhibit significantly higher errors in predicting the apex points across all bounces. This indicates that relying solely on rotational or translational data augmentation is insufficient to capture the roto-translational invariant properties required for accurate dynamics modeling. Typically, our proposed MNN achieves the lowest RMSE in predicting the apex between \#1-\#2 and \#2-\#3, indicating that MNN not only accurately predicts the trajectory after the bounce but also maintains the state more precisely compared to all other benchmarks. The state maintenance property will be explored in future work, particularly in scenarios where the ball makes contact with multiple surfaces, such as during interactions between the ball and the table or the ball and a paddle.


\begin{table}[h]
\caption{Comparison of Apex RMSE between bounces (in mm)}
\label{tab: apex rmse}
\centering
\begin{tabular}{l|cc|cc|cc}
\hline
\multirow{2}{*}{Method} & \multicolumn{2}{c|}{\textbf{Before \#1}} & \multicolumn{2}{c|}{\textbf{\#1 - \#2}} & \multicolumn{2}{c}{\textbf{\#2 - \#3}} \\ \cline{2-7} 
                        & \textbf{Mean}       & \textbf{Std}       & \textbf{Mean}       & \textbf{Std}      & \textbf{Mean}      & \textbf{Std}      \\ \hline
LSTM+Aug.               & 134.4               & 65.6               & 173.3               & 123.6             & 251.2              & 136.0             \\
Diffusion              & 47.9               &  56.8            & 58.4              & 64.1             & 96.3            & 103.5             \\
A-Tune+Aug.             & 17.6                & 19.7               & 56.7                & 32.0              & 118.6              & 34.1              \\
MLP+Aug.                & 20.5                & 11.5               & 58.1                & 23.5              & 108.3              & 52.9              \\
MLP+GS                  & 14.9                & 23.6               & 43.6                & 32.1              & 78.4               & 59.5              \\
Skip+GS                 & \textbf{14.6}       & 21.3               & 42.4       & 38.2              & 85.1               & 55.9              \\
MNN+GS                  &  14.8      & 25.1               & \textbf{37.2}              & 29.6              & \textbf{71.5}      & 54.9              \\ \hline
\end{tabular}
\end{table}

\section{LIMITATIONS AND FUTURE WORK}
The key limitation of our approach is its reliance on launcher settings for spin labeling which limit data collection to specific launchers. Additionally, using the GS process to extract roto-translational invariant representations may overly constrain the problem in cases where the ball contacts irregular surfaces or when dealing with non-spherical objects, such as an American football, which are beyond the scope of this paper. Moreover, in real gameplay, the ball is launched from a paddle rather than a ball launcher, and the dynamics of paddle-ball interaction are not captured in our current model. Lastly, integrating a factor graph into the learning framework significantly increases the training time. The inference time for 80 observations using an Nvidia RTX 3090 GPU is 18 seconds during training, and the autoregression time for 300 points is 0.046 seconds. However, the inference and prediction times could be significantly reduced by porting the model from a general-purpose deep learning framework to a compiled language such as C++, given the small model size (7346 parameters). Additionally, the factor graph inference speed could be greatly improved by using incremental algorithms such as ISAM2 \cite{kaess2012isam2}. Future work will focus on addressing these limitations and reducing computational times.
\section{CONCLUSIONS}
In this paper, we propose a novel dynamics learning framework, which enables the joint training of a estimator and a dynamics model.  We also leverage the Gram-Schmidt process to extract roto-translational invariant representations to largely improve the accuracy and generalization of our model. Additionally, we introduce a self-multiplicative neural network architecture to further enhance the nonlinearity and performance of the model, allowing it to capture the complex aerodynamics and bounce dynamics of the ball. Our approach outperforms the benchmarks listed in the results.







\bibliographystyle{IEEEtran}
\bibliography{References}

\end{document}